\PassOptionsToPackage{letterpaper, margin=1in}{geometry}
\documentclass[manuscript,screen,nonacm]{acmart}

\AtBeginDocument{%
  }

\usepackage{fdiaz}
\usepackage{color}
\usepackage{subfigure}
\usepackage{tcolorbox}
\usepackage{multirow}
\usepackage{adjustbox}
\usepackage{wrapfig}
\newcommand{\prob}{\mathop{\textrm{Pr}}}
\newcommand{\expect}{\mathop{\mathbb{E}}}

\newcommand{\instance}{x}
\newcommand{\inputSpace}{\mathcal{X}}
\newcommand{\target}{y}
\newcommand{\outputSpace}{\mathcal{Y}}

\newcommand{\loss}{\mathcal{L}}
\newcommand{\rmodel}[1]{g_{#1}}
\newcommand{\mlmodel}[1]{f_{#1}}
\newcommand{\rparam}[0]{\omega}
\newcommand{\mlparam}[0]{\theta}
\newcommand{\jointModel}[0]{\mathcal{M}}

\newcommand{\doc}{d}
\newcommand{\docSpace}{\mathcal{D}}
\newcommand{\docKeySpace}{\docSpace_{\text{key}}}

\newcommand{\query}{q}
\newcommand{\querySpace}{\mathcal{Q}}
\newcommand{\results}{r}
\newcommand{\resultSpace}{\mathcal{R}}
\newcommand{\feedback}{s}
\newcommand{\feedbackSpace}{\mathcal{S}}

\newcommand{\metric}{\mu}

\newcommand{\collection}{\textrm{C}}
\newcommand{\data}{\textrm{L}}
\newcommand{\udata}{\textrm{U}}

\newcommand{\testdata}{\data_{\text{test}}}

\newcommand{\QuerySet}{Q}

\begin{document}

\title{Retrieval-Enhanced Machine Learning: Synthesis and Opportunities}

\author{To Eun Kim}
\orcid{0000-0002-2807-1623}
\affiliation{%
  \institution{Carnegie Mellon University}
  \state{PA}
  \country{United States}
  \postcode{15213}
}
\email{toeunk@cs.cmu.edu}

\author{Alireza Salemi}
\orcid{0009-0006-1937-2615}
\affiliation{%
  \institution{University of Massachusetts Amherst}
  \state{MA}
  \country{United States}
  \postcode{01003}
}
\email{asalemi@cs.umass.edu}

\author{Andrew Drozdov}
\authornote{Now at Databricks.}
\orcid{0000-0002-1025-5715}
\affiliation{%
  \institution{University of Massachusetts Amherst}
  \state{MA}
  \country{United States}
  \postcode{01003}
}
\email{adrozdov@cs.umass.edu}

\author{Fernando Diaz}
\orcid{0000-0003-2345-1288}
\affiliation{%
  \institution{Carnegie Mellon University}
  \state{PA}
  \country{United States}
  \postcode{15213}
}
\email{diazf@acm.org}

\author{Hamed Zamani}
\orcid{0000-0002-0800-3340}
\affiliation{%
  \institution{University of Massachusetts Amherst}
  \state{MA}
  \country{United States}
  \postcode{01003}
}
\email{zamani@cs.umass.edu}


\begin{abstract}
In the field of language modeling, models augmented with retrieval components have emerged as a promising solution to address several challenges faced in the natural language processing (NLP) field, including knowledge grounding, interpretability, and scalability.
Despite the primary focus on NLP, we posit that the paradigm of retrieval-enhancement can be extended to a broader spectrum of machine learning (ML) such as computer vision, time series prediction, and computational biology.
Therefore, this work introduces a formal framework of this paradigm, Retrieval-Enhanced Machine Learning (REML), by synthesizing the literature in various domains in ML with consistent notations which is missing from the current literature.
Also, we found that while a number of studies employ retrieval components to augment their models, there is a lack of integration with foundational Information Retrieval (IR) research.
We bridge this gap between the seminal IR research and contemporary REML studies by investigating each component that comprises the REML framework.
Ultimately, the goal of this work is to equip researchers across various disciplines with a comprehensive, formally structured framework of retrieval-enhanced models, thereby fostering interdisciplinary future research.
\end{abstract}




\maketitle

\section{Introduction}
\label{sec:introduction}

\paragraph{\textbf{Background}}
In recent years, the research landscape surrounding large language models (LLMs) has witnessed substantial growth, underscored by the profound potential these models hold for various natural language processing (NLP) tasks. One of the significant advancements that has propelled this field forward is the scaling of the number of parameters of LLMs, which has enabled the training of models with unprecedented size and complexity \cite{zhao2023surveyLLM}. We witness a similar trend in other fields adjacent to machine learning, for example, large vision foundation models for representing images and videos \cite{dosovitskiy2021an,Arnab_2021_ICCV}. Concurrently, the notion of in-context learning (ICL) \cite{dong2022surveyICL} has emerged as a transformative capability, allowing LLMs to dynamically adapt and incorporate new information during its inference.
In parallel, the information retrieval (IR) community has been actively exploring techniques aimed at improving the efficiency, effectiveness, and robustness of accessing information from large-scale collections.

The convergence of these two domains has given rise to a new trend in research, where models 
are equipped with retrieval capabilities to access external knowledge during both training and inference stages \cite{mialon2023-ALM-survey,zamani:reml}. This integration of retrieval mechanisms into the prediction pipeline started to gain significant traction, as it allows models to ground their predictions in external knowledge without necessitating an increase in model capacity. Methods presented by \citet{Hashemi2020GT} and \citet{Lewis+al:2020} are among the earliest work in this space; the former focuses on retrieval-augmented representation learning by extending the transformer network, while the latter studies the paradigm of retrieval-augmented generation (RAG) for knowledge-intensive language tasks. That  said, using retrieval results to improve a machine learning systems is not new. Pseudo-relevance feedback methods---methods for representing search queries using the top retrieved documents---are perhaps the first set of methods in this category \cite{Attar:1978,Croft:1979}. 
The ICL ability inherent in the LLMs has played a pivotal role in facilitating the dissemination and adoption of these retrieval-augmented approaches. By integrating retrieved documents into the prompt of the LLMs, researchers have been able to harness the external knowledge sources without fundamentally altering the underlying model architecture.

\paragraph{\textbf{Motivation}}
Since improving model performance by increasing the number of parameters is not sustainable, one motivation of retrieval-based models stems from the finding that, while large models tend to memorize training data \cite{carlini2021extracting}, incorporating retrieval-based methods can effectively transfer the burden of memorization to external storage systems \cite{borgeaud:retro, shao2024scalingretrievalbasedlanguagemodels}. 
We advocate for enhancing machine learning (ML) models in general (i.e., beyond generation) with the ability to employ stored information via information retrieval techniques. IR has already shown its merits in aiding human interaction with vast text databases. We posit that IR's utility can be broadened to enable machine access to not only extensive text databases but also to knowledge represented in more abstract forms. 
By integrating ML architectures with direct access to IR systems, we aim to separate the processes of reasoning and memory. 
\citet{zamani:reml} dubbed this approach, \emph{retrieval-enhanced machine learning (REML)}, as a broader concept that extends ML. Extending their work, we further survey the recent advances of REML in the field of ML, including NLP, with consistent mathematical notation. By doing so, we aim to equip researchers with a comprehensive and structured overview of the REML methodologies, enabling them to swiftly embark on research within this domain.

\paragraph{\textbf{Applications of REML}}
The landscape of REML paradigm encompasses a diverse array of sub-domains, each with its unique set of challenges and applications. This includes seminal work in language modeling \cite{guu-realm,Lewis+al:2020,borgeaud:retro,izacard-grave-2021-leveraging,zhong-etal-2022-training,izacard_few-shot_2022,ram2023incontext,wang2023visuallyaugmented,Shi2023REPLUGRB,li2022decoupled,Khandelwal2020Generalization},  machine translation \cite{khandelwal2021-knn-mt}, question answering \cite{KG-FiD, chen-etal-2017-reading,lee-etal-2019-latent,nakano2022webgpt,lazaridou2022internetaugmented, wu-etal-2022-efficient, qamat, zhang2024retrievalqaassessingadaptiveretrievalaugmented, kim-etal-2023-tree}, fact verification \cite{Lewis+al:2020, petroni2023improving, chen2023complex},  open domain \cite{shuster-etal-2022-language,komeili:blenderbot2,thoppilan2022lamda} and task-oriented \cite{thulkeDSTC2021,eric-etal-2017-key,raghu-etal-2021-end, nekvinda2022:aargh-tod} dialog systems, slot filling \cite{glass:zeroshot-slotfilling}, state tracking \cite{king2023:ra-dst}, reinforcement learning  \cite{fernandez:policy-reuse,goyal:rarl,humphreys:large-scale-rarl, chowdhury2024retrievalguided},  computer vision \cite{Chen2022ReImagenRT, yasunaga2023:multimodal-LM, ramos2023:lmcap-imagecaptioning, Shrestha_2024_CVPR}, multimodal ML \cite{kif, iscen2024retrievalenhanced}, commonsense reasoning \cite{yu2022:commonsense-unified}, evidence attribution \cite{gao:alce, menick:deepmind-gophercite, huo2023:remlworkshop-evidence, gao:rarr}, 
knowledge-graph augmentation 
\cite{baek-etal-2023-knowledge, ju-etal-2022-grape, knowledge-enhanced-plm, zhang-etal-2022-subgraph-enhanced, KG-FiD, wei2024llmrec}, 
personalization \cite{salemi:lamp, salemi2024optimization},  mathematical problem-solving \cite{yang:leandojo}, code generation \cite{zhang-etal-2023-repocoder, zhou2023docprompting, wang2024coderagbenchretrievalaugmentcode}, representation learning for audio and speech \cite{sanabria2023acoustic, lin2024speechdpr, zhouICASSP24asr}, time series prediction \cite{jing2022timeseries, yang2022mqretnn, xu2024rebar}, robot navigation and embodiment \cite{xie2024embodiedrag, anwar2024remembrbuildingreasoning}, chip design \cite{chowdhury2024retrievalguided}, medical reasoning \cite{medicalReasoning}, drug discovery \cite{liu2024conversational}, and protein structure prediction \cite{jumper2021alphafold, cramer2021alphafold2}. 
The industry and open source communities have swiftly embraced the adoption of retrieval-based models, recognizing their potential for accelerated adaptation and performance enhancement. Frameworks such as LangChain,\footnote{https://www.langchain.com} LlamaIndex,\footnote{https://www.llamaindex.ai} and DSPy \cite{khattab2024dspy} have emerged, streamlining the process of implementing the retrieval-based models. 
This broad spectrum of domains (not an exhaustive list) underscores the versatility and impact of the REML paradigm across diverse applications.

\paragraph{\textbf{Main Contributions of This Work}}
Although many current applications are centered around natural language processing, we believe that ML models that leverage retrieval components are not confined to language models alone, but can be extended to any ML models. To address this broader applicability, we formalize the framework as Retrieval-Enhanced Machine Learning (REML) and synthesize existing studies with consistent mathematical notations which is lacking in the current literature. Throughout this paper, we consistently use the notations described in Table \ref{tab:notations}.
Also, despite the advancements in REML models, there remains a significant underutilization of the rich and extensive body of work from information retrieval research which can offer numerous methodologies and insights that can substantially benefit REML models. This work aims to bridge this gap by integrating IR research into the design of REML models.
Ultimately, we hope this work will enable researchers across various fields leveraging ML to easily understand the framework of REML and its extensibility.

\section{Retrieval-Enhanced Machine Learning}
\label{sec:overview}

\begin{table}[t]
    \centering
    \adjustbox{max width=\textwidth}{
    \begin{tabular}{lllll}\toprule
        Notation & Description && Notation & Description \\\cline{1-2}\cline{4-5}
        $\instance$ & input instance && $\doc$ & retrieval item \\
        $\inputSpace$ & input space && $\docSpace$ & retrieval space \\
        $\target$ & output target && $\docKeySpace$ & retrieval key space \\
        $\outputSpace$ & output space && $\collection$ & stored retrievable items \\
        & && $\query$ & query \\
        $\data$ & labeled data (i.e., $\data \subset \mathcal{X} \times \mathcal{Y}$) && $\querySpace$ & query space \\
        $\udata$ & unlabeled data (i.e., $\udata \subset \mathcal{X} $) && $\results$ & retrieval results \\
        &&&$\resultSpace$ & retrieval result space \\
        $\loss$ & The downstream loss function && $\feedback$ & model feedback \\
        $\mlmodel{\mlparam}$ & a predictive machine learning model parameterized by $\theta$ && $\feedbackSpace$ & model feedback space \\
        $\rmodel{\rparam}$ & a retrieval model parameterized by $\omega$ && $\metric$ & evaluation metric \\
        $\jointModel$ & a tuple of predictive and retrieval model \\      
        \bottomrule
    \end{tabular}}
    \caption{Notations used in this paper to synthesize REML research.}
    \label{tab:notations}
    \vspace{-0.7cm}
\end{table}




To begin an in-depth exploration of retrieval-enhanced machine learning (REML), we commence with reiterating the generalized formal definition of the task set by \citet{zamani:reml}. Like all predictive machine learning frameworks, subsequently referred to as ML models, REML is tasked with learning a functional relationship that maps an input space $\mathcal{X}$ to an output space $\mathcal{Y}$. Unlike other ML models, REML predicts outcomes through interactions with one or more information access models, each facilitating access to a database or repository of knowledge. 
Hence, REML is formally articulated as $y = f_{\theta}(x; g_{\omega_1}, g_{\omega_2}, \cdots, g_{\omega_N})$, where $x \in \mathcal{X}$ and $y \in \mathcal{Y}$ symbolize the input instance and target output respectively, $\mlmodel{\mlparam}$ represents an ML model parameterized by $\theta$, and $\rmodel{\rparam_i}$ represents an information access model parameterized by $\omega_i$. Here, $N$ signifies the total number of information access models that $f_\theta$ can consult with. Each $\rmodel{\rparam_i}$ is associated with a collection, repository, or memory $\collection_i$, which might consist of natural language texts or alternative indexed representations. Consequently, the collection $\collection_i$ serves as a versatile array of parameters available to $\rmodel{\rparam_i}$, which may be employed \textit{ad hoc}, in the same way as many non-parametric and lazy learning techniques. 
\citet{zamani:reml} outlines three \textit{necessary} (Reqs) and \textit{optional} (Opts) requirements for REML models.


    
    

    

\begin{enumerate}
\item[Req 1] \textbf{Querying}. Every $\mlmodel{\mlparam}$ should possess a capability to generate queries that are dependent on the input, directed towards $\rmodel{\rparam_i}$s.
Refer to \S\ref{sec:querying}.

\item[Req 2] \textbf{Retrieval}. Every $\rmodel{\rparam_i}$ must be capable of processing the queries from $\mlmodel{\mlparam}$, fetching pertinent information from its corresponding repository $\collection_i$.
Refer to \S\ref{sec:searching}.

\item[Req 3] \textbf{Response Utilization}. Every $\mlmodel{\mlparam}$ should utilize the information obtained from $\rmodel{\rparam_i}$s in its prediction-making process.
Refer to \S\ref{sec:presentation}.

\item[Opt 1] \textbf{Storing}. $\mlmodel{\mlparam}$ may archive some information in  $\collection_i$ for later retrieval, applicable during both training and inference.
Refer to \S\ref{sec:storing}.

\item[Opt 2] \textbf{Feedback}. $\mlmodel{\mlparam}$ may have the functionality to offer feedback to $\rmodel{\rparam_i}$, during the training and inference for improvements of either $\mlmodel{\mlparam}$, $\rmodel{\rparam_i}$, or both.
Refer to \S\ref{sec:optimization}.
\end{enumerate}

\begin{figure*}[t]
    \centering
    \subfigure[Retrieval-only]{\label{fig:reml:a}        \includegraphics[trim={4cm 7.5cm 13cm 0cm},clip,width=.24\textwidth]{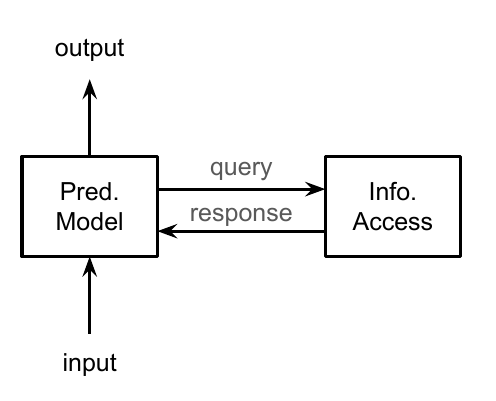}}
    \subfigure[Retrieval with memory]{\label{fig:reml:b}        \includegraphics[trim={13cm 7.5cm 4cm 0cm},clip,width=.24\textwidth]{02-overview/graphics/REML.pdf}}
    \subfigure[Retrieval with feedback]{\label{fig:reml:c}        \includegraphics[trim={4cm 0cm 13cm 7.5cm},clip,width=.24\textwidth]{02-overview/graphics/REML.pdf}}
    \subfigure[Retrieval with memory \& feedback]{\label{fig:reml:d}        \includegraphics[trim={13.5cm 0cm 3.5cm 7.5cm},clip,width=.24\textwidth]{02-overview/graphics/REML.pdf}}
    \vspace{-0.5cm}
    \caption{Retrieval-enhanced machine learning models should implement three necessary requirements (querying, retrieval, and response utilization) and may implement two optional properties (storing information and providing feedback to the information access model). This results in four categories of REML models presented above. Figure is taken from \cite{zamani:reml}.}
    \label{fig:reml}
    \vspace{-0.5cm}
\end{figure*}

The simplest form of REML model is depicted in \figurename~\ref{fig:reml:a} by focusing solely on the essential criteria. The second category, illustrated in \figurename~\ref{fig:reml:b}, utilizes the first optional property by storing information in a storage for subsequent retrieval. The third category, presented in \figurename~\ref{fig:reml:c}, employs the second optional property, offering feedback to the information access models. The final category incorporates all optional properties, as detailed in \figurename~\ref{fig:reml:d}.

\begin{figure}[t]
    \centering
    \includegraphics[width=0.55\linewidth]{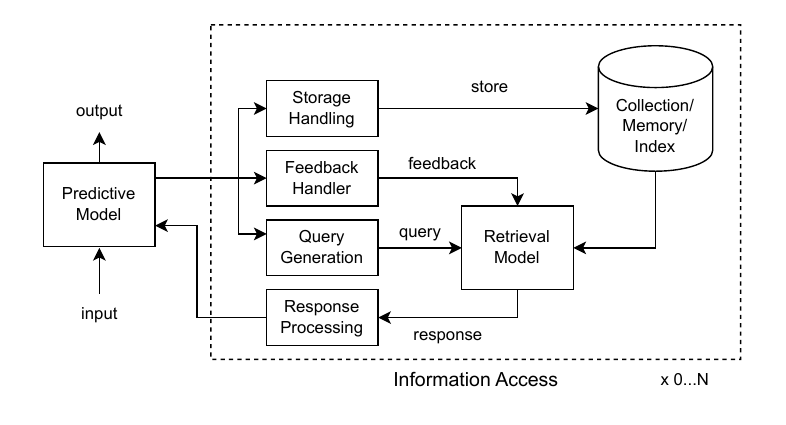}
    \vspace{-0.5cm}
    \caption{A generic framework for REML \cite{zamani:reml}.
    The multiplicative nature of the information access process implies that the access to the information can be distributed and/or be done iteratively.
    Note that each component do not have to be completely separated, e.g., Query Generation or Response Processing module can be dealt within the Predictive Model. In abstract, however, we consider them as one of the components of information access process that can be described separately.}
    \vspace{-0.4cm}
    \label{fig:reml-framework}
\end{figure}

Based on the requirements, \citet{zamani:reml} proposed a comprehensive framework for REML, as illustrated in \figurename~\ref{fig:reml-framework}. This framework is structured around two principal components: the prediction model $\mlmodel{\mlparam}$ and the information retrieval models $\rmodel{\rparam_i}$s. For any given input $\instance$, the predictive model $\mlmodel{\mlparam}$ has the flexibility to initiate multiple retrieval operations. This could involve dispatching multiple queries, engaging with numerous data repositories, offering feedback to the information retrieval components, or employing a mix of these strategies. It's noteworthy that for certain inputs, the number of retrieval processes might be nil, thereby allowing REML to extend the conventional predictive modeling.

\section{Querying}
\label{sec:querying}

In REML, the process of acquiring information hinges upon the act of querying a knowledge or information repository. Consequently, the formulation of a query from the input, whether unstructured or structured, stands as the pivotal initiation point for the interplay between predictive and retrieval models within the REML framework. The following sections introduce common operations employed to generate queries based on the task's input.

\subsection{Deciding Where to Query}
Before sending a query to an information access system, a REML system can decide where its query should be sent to.
Bearing in mind that each $\rmodel{\rparam_i}$ is associated with $\collection_i$ (multiplicative nature of information access process depicted in Figure \ref{fig:reml-framework}), the Query Generation module first decides which tuple(s) of $\rmodel{\rparam_i}$ and $\collection_i$ should be selected depending on the context which aligns with the mixture-of-expert-like interpretation described by \citet{pan2023knowledgeincontext} \footnote{In KiC \cite{pan2023knowledgeincontext}, a router selects an expert predictive model (not a retriever) with specific knowledge source. However, this helps the understanding of the first step of the Query Generation module.}.
This \textit{query decision} problem can be understood by the following sub-problems. 1) \textit{Corpus Selection}: deciding which corpora are needed to be searched over (can be null when no retrieval is needed), and 2) \textit{Retriever Selection}: deciding which retrieval model should be used to search over the chosen corpora.

\subsubsection{Corpus Selection}
Selecting $\collection_i$ is not only regarded with deciding what kind of external information should be provided to the $\mlmodel{\mlparam}$ \cite{pan2023knowledgeincontext} but also encompasses the question of \textit{when to query} (no corpus selection when no retrieval is beneficial to $\mlmodel{\mlparam}$).
This can be a critical question as retrieval augmentation can hurt the performance for certain input types \cite{maekawa2024retrieval, mallen-etal-2023-when-not-to-trust, asai2024selfrag}. It also can save computational resources by reducing the number of searching \cite{labruna2024retrieve}.
There can be several criteria on whether external information can be beneficial to the predictive model. It can be based on term popularity \cite{mallen-etal-2023-when-not-to-trust}, input complexity \cite{jeong2024adaptiverag}, or a trained model \cite{asai2024selfrag}.

\subsubsection{Retriever Selection}
Once one or more $\collection_i$ is selected, a REML system can further benefit itself by choosing the optimal retriever specialized in searching the selected corpora. This is a relatively challenging and new task, and we refer readers to \citet{khramtsova2023selectingwhichdense, khramtsova2024leveraging} for deeper understanding.

\subsection{Reformulating the Input} 
For many cases, raw user input cannot be directly leveraged as a query to the retrieval model, underscoring the critical need for input reformulation into a different representation. This reformulation occurs through a process, where the input is transformed using a separate component or the same predictive model based on the specific requirements of the system. The general equation for transformation is
\begin{equation}
    \query = \textit{transform}_q(\instance, context)
\end{equation}
\noindent
where $\instance$ is the input that the transformation should be applied on (i.e., the original input to the task, the previous search query, the output of another transformation, etc.) and $context$ is side information that this function can use in performing the transformation. For example, one of the use cases of $context$ can be a user profile, which can help personalize this transformation for a user \cite{salemi:lamp}. Common motivations for transforming the input into an alternative format encompass a range of factors, including but not limited to truncation, expansion, and conversion.

\subsubsection{Compression} In certain scenarios, not all words or components of the input prove relevant for the search objective. Consequently, the common practice of omitting specific segments of the input has been employed in numerous prior studies. In the majority of cases, sequence-to-sequence models are trained to identify and mark the segments that require reduction \cite{ni-etal-2019-learning, musa2019answering, 10.1145/3477495.3531793, khashabi-etal-2017-learning}. At times, a straightforward approach such as segmenting the input into distinct chunks and utilizing these segments as queries can be highly effective \cite{borgeaud:retro}. In multi-modal search scenarios, omitting a specific modality from the input and conducting searches across a corpus from different modalities has proven to be valuable \cite{kat}.

\subsubsection{Expansion} In certain scenarios, the input alone may lack essential information required by the search system to yield desired results. In such situations, augmenting the input with additional pertinent data can be done. This process of expansion broadens the context and enhances the search system's capability to retrieve relevant and meaningful results, thereby improving overall system performance. Typically, expansion is achieved by concatenating the input with previously retrieved results \cite{zhu-etal-2021-adaptive, xiong2021answering} or generated text that is conditioned on the input \cite{wang2023query2doc, mao-etal-2021-generation, liu2022query, chuang-etal-2023-expand}.

\subsubsection{Conversion} For some situations, reshaping the input into a new query based on its inherent structure, instead of mere expansion, is proved to be advantageous. This approach is particularly valuable when crafting structured queries for database \cite{arcadinho-etal-2022-t5ql, dou2022unisar} and API access \cite{schick2023toolformer, qin2023toolllm, NEURIPS2022_b1efde53, jin2023genegpt}. 

The conversion operation may results in a transformation of the input space $\inputSpace$ into the query space $\querySpace$. Consequently, it is essential to note that $\querySpace$ and $\inputSpace$ are not necessarily equivalent, signifying that the transformed queries might operate in a distinct space compared to the original input. In complex multi-modal search scenarios, employing the input directly might not be viable. Consequently, transforming the input into a different modality form becomes imperative, ensuring seamless and efficient communication between the predictive model and the search system \cite{gao2022thousand, lin-byrne-2022-retrieval, wu-mooney-2022-entity, lin2023finegrained, dedr}. 

Moreover, transforming the input from the original input space into the latent space of a language model and retrieving information from the model's prior interactions with data, like what happens in kNN-LM \cite{Khandelwal2020Generalization}, represent additional transformations that alter the query space \cite{NEURIPS2022_97011c64, yogatama-adaptive-semiparametric, he-etal-2021-efficient, khandelwal2021-knn-mt, kassner-schutze-2020-bert}. Indeed, Neural Turing Machines \cite{graves2014neural, gulcehre2017memory, Rae2016-scaling-ntm} and Memory Transformers \cite{zhong-etal-2022-training, wu-etal-2022-memformer, wan-etal-2022-g} employ a similar conversion process to translate input into a latent variable. This transformation is essential for enabling effective retrieval from the memory/storage component of these models.

\subsection{Decomposing the Input}

This category involves breaking down a complex input into simpler parts, often to better understand the content and retrieve more accurate results. This technique is particularly useful when dealing with long and complex inputs that cover multiple topics or concepts \cite{min-etal-2019-multi, perez-etal-2020-unsupervised, zhou-etal-2022-learning-decompose}. The general equation for decomposition is
\begin{equation}
    \QuerySet = \textit{decompose}(\instance, context)
\end{equation}
\noindent
where $\instance$ is the input that should be decomposed (i.e., the original input to the task, the output of a transformation, etc.), $context$ is side information that this function can use in performing the decomposition, and $Q$ is the set of decomposed queries. Note that decomposition returns a set of queries in contrast with the transformation operation which only returns a single query.

\subsection{Unified Equation for Query Generation} The unified equation for query generation is therefore
\begin{equation}\label{eq:querying-unified}
    \QuerySet = \textit{decompose}(\textit{transform}_q(\instance, context), context)
\end{equation}

Any of these function can be replaced with the identity function \cite{izacard-grave-2021-leveraging, karpukhin-etal-2020-dense,asai-etal-2022-evidentiality,yamada-etal-2021-efficient, Lewis+al:2020, Lewis+al:2020, thorne-etal-2018-fever, guu-realm} disregarding modality \cite{Salemi2023PreTrainingMD, qu-2021-multimodal-passage-retrieval}. Furthermore, it is conceivable to apply these functions multiple times and in various order. Particularly in the realm of multi-hop question answering and fact verification, prior research extensively employs multiple transformations and decompositions to fulfill the task requirements \cite{qi-etal-2019-answering, trivedi-etal-2023-interleaving, yadav-etal-2020-unsupervised, das-etal-2019-multi, jiang2023active-flare}. Given the intricacies of these tasks, leveraging a combination of various transforms and decompositions becomes essential.

\section{Searching}
\label{sec:searching}

Depending on the nature of the documents, the queries, and the tasks, different search functionalities are required and expected. For instance, in some task-oriented dialogue systems, $\mlmodel{\mlparam}$ (conversational agent) requires access to relational databases. Therefore, in such scenarios, structured queries like SQL is used for searching.
That being said, retrieval items in most applications are in the form of semi-structured or unstructured text or involve multi-modal aspects. In the following, we review different retrieval models for the various situations.

\subsection{Retrieval Models with Sparse Representations} 
Many text-based retrieval models use sparse representations for representing queries and documents. For instance, term-matching (lexical) retrieval models, such as TF-IDF \cite{Salton1988TFIDF}, BM25 \cite{Robertson1995OkapiBM25}, and query likelihood \cite{Ponte:1998}, represent each query and document using a $V$-dimensional sparse vector, where $V$ denotes the vocabulary size. In these models, the dimensions associated with the terms that appear in the given text carry non-zero values and the rest are zero. Most of these models are based on the term independent, or the bag-of-words assumption. That being said, models that consider term position and ordering exist, such as higher-order language models \cite{Song:1999}, positional language models \cite{Lv:2009}, and sequential dependency models \cite{Metzler:2005}. 

The sparsity nature of data in these retrieval models enable them to use an inverted index data structure for scalable and efficient retrieval. Note that these models often suffer from a vocabulary mismatch problem, meaning that using different vocabulary for representing the same concept in the query and document does not contribute to the estimated relevance score. This can significantly impact the performance of $\rmodel{\rparam}$, especially from the recall perspective. Query expansion and document expansion approaches exist to address the vocabulary mismatch problem, including the pseudo-relevance feedback models \cite{Lavrenko:2001:RM,Zhai2001Mix,Rocchio:1971}. Neural network solutions for expanding the documents, such as SPLADE \cite{splade}, has shown promising results when sufficiently large-scale training data is available.

An alternative to lexical representation is using latent vectors. For instance, SNRM \cite{zamani-snrm} learns high-dimensional sparse latent vectors produced by deep learning models for representing queries and documents.

\subsection{Retrieval Models with Dense Representations}
Queries and documents can be represented using low-dimensional (compared to the vocabulary size) dense vectors. Such dense vectors are often obtained using pre-trained language models, such as BERT \cite{devlin-etal-2019-bert}, that are fine-tuned for retrieval tasks \cite{karpukhin-etal-2020-dense}. Dense retrieval models are commonly based on bi-encoder architectures -- one encoder for representing the query and another for the document. These encoders could share parameters. Some dense retrieval methods, such as DPR \cite{karpukhin-etal-2020-dense}, represent each query or document by a single dense vector. While others, such as ColBERT \cite{Khattab:2020:colbert}, use one vector per token, resulting in multiple vectors for each query and document. Approximate nearest neighbor (ANN) algorithms, such as HNSW \cite{hnsw}, are used for efficient retrieval when dealing with dense representations. Dense retrieval approaches are also commonly used when dealing with multi-media and multi-modal data \cite{qu-2021-multimodal-passage-retrieval,dedr}.

\subsection{Reranking Models}
Modern search engines are mainly designed based on a multi-stage cascaded architecture--a stack of ranking models where the first model efficiently retrieves a list of documents and the following models rerank the results from the previous stage. A common scenario is a two-stage process: retrieval and reranking. Reranking models are often optimized using explicit or implicit relevance labels. These models are called learning-to-rank models. Early learning-to-rank models rely on manually-extracted and engineered features sets, while the most recent ones rely on deep learning models for representation learning and reranking. A common strategy for reranking using deep learning models is called cross encoding \cite{nogueira2019bert-rerank}, meaning that a query and a candidate document are concatenated and fed to a network like BERT \cite{devlin-etal-2019-bert}, which is trained (or fine-tuned) to produce a relevance score. Learning-to-rank models can be optimized using pointwise, pairwise, or listwise loss functions. For more information, refer to the learning-to-rank survey by \citet{Liu2009LTR} and the neural ranking model surveys by \citet{Mitra:2018:NeuralIR} and \citet{Guo:2020:NeuralIR}.

\subsection{Generative Retrieval Models}
Generative retrieval models, or differentiable search indexes, adopt an encoder-decoder or a decoder-only neural network architecture with the goal of generating document identifiers given the query. Even though earlier attempts to developing these models \cite{tay2022transformer} fail at performing effectively at scale \cite{pradeep-etal-2023-generative}, recent research by \citet{Zeng:2024:GR} developed prefix-oriented optimization approaches that let generative retrieval models to effectively scale up to large collections. These models often assign a semantic document identifier to each document in the collection and are optimized to generate the identifiers of relevant documents using sequential decoding algorithms, such as beam search.

\subsection{Unified Equation for Searching}\label{subsec:search-eq}

The current literature of searching can be generalized by equations of \textit{addressing} and \textit{reading} borrowed from the paradigm of Neural Turing Machines (NTM) \cite{graves2014neural, graves2016hybrid, Rae2016-scaling-ntm}. 
Before reading from the collection $\collection_i$, the model should decide which part of the collection it should attend to by addressing. The addressing can be done by comparing the query $\query_t \in \QuerySet$ (from \ref{eq:querying-unified}) with the keys in the collection and/or by finding the location in the collection.
The addressing is also used when constructing the collection which will be discussed in Section \ref{sec:storing}.

\subsubsection{Content-based addressing}
At $t$'th iteration, with a slight simplification of the notation (simplifying $\collection_i^t$ to $\collection_t$), given a query $\query_t$ and a collection $\collection_t$, content-based addressing can be defined as:
\begin{equation}\label{eq:content-addressing}
    w_t^{content} = \textit{address}_\textit{content}(\query_t, \collection_t) =
    topK(sort(\textit{score}(\query_t, \textit{transform}_s(\collection_t))), k)
\end{equation}
\noindent
where \textit{k} is the number of relevant addresses to be selected based on the query, \textit{score} is a scoring function, such as BM25 \cite{Robertson1995OkapiBM25} or cosine similarity \cite{guu-realm, majumder2023clin}. The content-based address vector $w_t^{content}$ can be exhaustively computed by pairwise comparisons of query and all elements of the collection \cite{graves2014neural, graves2016hybrid, weston2015memory-networks, metalearning-mann, seqrec-user-memnet2018, sukhbaatar2015end}, accelerated \cite{guillaum-19-large-memory-product-keys, weston2015memory-networks}, or approximated by ANN, selecting the top $k$ items $\doc_i\in\collection$, resulting in $w_t^{content}$ with $k$ non-zero elements \cite{Rae2016-scaling-ntm, kumar2016dynamicmemory, wu2022memorizingtransformers, Khandelwal2020Generalization, zhong-etal-2022-training, alon2022neurosymbolic, majumder2023clin}.
The function $\textit{transform}_s$ may be needed when the contents in the collection cannot be readily compared with the $\query_t$, e.g., mapping to a feature space \cite{weston2015memory-networks, Shi2023REPLUGRB, guu-realm, borgeaud:retro} and lexical transformation \cite{madaan-etal-2022-memprompt}. This function can be an identity function when transformation is not needed \cite{grave2017continuous-cache}.

\subsubsection{Location-based addressing}
Location-based addressing lets the searching system access the corpus purely by storage location such as index without any lexical or contextual comparison between a query and the corpus. Therefore, this is often used for storing (Section \ref{sec:storing}) or recency-based retrieval. Thus, 
\begin{equation}\label{eq:location-addressing}
    w_t^{location} = \textit{address}_\textit{location}(\query_t, context)
\end{equation}
\noindent
where \textit{context} is the side information that this function can use in generating the location based address (e.g., previous generated addresses by this function).
For some applications, both content- and location-based addressing can be used together. To this end, the final address can be defined as:
\begin{equation}\label{eq:searching-beta}
    w_t = {combine}(w_t^{location}, w_t^{content})
\end{equation}
\noindent
where \textit{combine} is a function that generates an address based on the location-based and content-based addresses. 
For example, most previous work relies on either purely content-based addressing \cite{metalearning-mann, guu-realm, Khandelwal2020Generalization, Rae2016-scaling-ntm} or purely location-based addressing \cite{weston2015memory-networks, shinn2023reflexion}. However, it is also possible to combine both content-based and location-based addressing \cite{graves2014neural, graves2016hybrid}.

\subsubsection{Unified Equation for Searching}
With the final address vector $w_t$, the retrieval results $\results_t$ from $\rmodel{\rparam}$ is defined as:
\begin{equation}\label{eq:searching-read}
    \results_t = \textit{read}(w_t, \textit{transform}_s(\collection_t)),
\end{equation}
where $\textit{read}$ simply selects the content from corpus in location $w_t$ that that is transformed by the $\textit{transform}_s$.

\section{Presentation \& Consumption}
\label{sec:presentation}

In this section we will cover two key parts of REML. \textbf{Presentation} involves not only how we define the result space $\resultSpace$, but also how the results from retrieval are prepared for the next step of consumption. Based on the application, the presentation stage can range from simple copying of the results to more complex pipelines with intermediate preprocessing and model-based transformations. 
\textbf{Consumption} is the process through which the predictive model ($\mlmodel{\theta}$) incorporates the retrieved information. 
There are many considerations when designing effective methods for presentation and consumption.
One typically wants to incorporate as much information as possible  while balancing the tradeoffs between efficiency and accuracy.

\subsection{Presentation}

When presenting search results to a human reader the interface is designed to make the findings easily consumed such as through sorting items by relevance or highlighting salient snippets \cite{white2003granular}. In REML, we follow a similar principle except the target consumer of the retrieved data is a machine, which has a different set of limitations and capabilities. 
Table \ref{tab:presentation} summarizes the research related to presentation.

\subsubsection{Transforming the data\label{sec:pres_transform}} Dependent on the task and source of data, the result data will be incomplete prior to consumption. The transformation of data is a general and powerful process that converts data through a separate model depending on the needs of the system.
Common reasons requiring further data transformation include decontextualization, translation, and summarization among others, thus $r$ can be transformed by the following equation:
\begin{equation}
    r' = transform_p(r)
\end{equation}

\paragraph{Decontextualization} When the retrieved item is only a few sentences of a much larger document, then it may require \textit{decontextualization} to resolve anaphora or previously defined abbreviation \citep{newman-etal-2023-question}.

\paragraph{Translation} It may be the case that search operates in a cross-lingual representation space, and there will be a mismatch in language between the retrieved items and desired output language \citep{10.1145/564376.564408}. A multilingual language model may be robust to code switching in the retrieved context, but it will likely be more reliable to translate any retrieved documents before processing them for prediction \cite{10.1145/1458082.1458179, jiang2024pretraining}. Translation can be applied to other modalities, such as regenerating an image to match an expected style.

\paragraph{Summarization} Due to possible context limits of predictive model, it is desirable to condense document data so that more documents can fit into the context. This can be achieved through automatic summarization, converting the original data into a shortened form through an extractive or abstractive process \citep{gao:alce, Li2023TeachLT, kim2024sure, xu2024recomp}. Furthermore, data can be summarized in the context of the input, providing clarity and explaining why the document is relevant.

\subsubsection{Combining result items\label{sec:pres_compose}} To further optimize the presentation of the result items for size or clarity, multiple items can be combined, e.g., summarizing all items jointly \citep{Wu2021RecursivelySB, Sarthi2024RAPTORRA}. Not all REML systems will combine items, as operating over individual items can be efficient and effective. Furthermore, combining items may lead to complications such as miscalibration between the individual scores and scores of a combined result, which is represented by \textit{compose}:
\begin{equation}
    r' = compose(r)
\end{equation}

\subsubsection{Truncate results list\label{sec:pres_truncate}} If not all documents fit into the context for consumption, we discard or truncate documents based on these limits, optimizing for length and potentially other properties such as diversity \citep{FiD-Light, Bahri2020ChoppyCT, meng2024ranked}, which is represented by \textit{truncate}:
\begin{equation}
    r' = truncate(r)
\end{equation}

\subsubsection{Unified Equation for Presentation} The full equation for presentation:
\begin{equation}
    r' = truncate(compose(transform_p(r)))
\end{equation}
\noindent
where any of these functions can be replaced with simple forms such as an identity function \citep{Lewis+al:2020, izacard-grave-2021-leveraging, izacard2021distilling}. Additionally, we can imagine these functions being applied multiple times in any order. We include this ordering as the one that seems most natural when taking into account context length limits. $transform_p$ is on individual items, and $compose$ is similar to $transform_p$ but on groups of items. Not shown is \textit{loading} of retrieved items. Unlike traditional ML systems, where data is typically used only for training, REML systems have unique requirements associated with data loading. Since the amount of external data required for an input is dynamic and considerable \cite{borgeaud:retro}, efficient load and manage of data is essential \cite{douze2024faiss, guo2020scann}. We assume loading is handled implicitly by the retrieval module.

\begin{table}[t]
    \centering
    \adjustbox{max width=\textwidth}{
    \begin{tabular}{l|p{12cm}}
    \toprule
        \toprule
         \multicolumn{2}{c}{Transform (\S\ref{sec:pres_transform})} \\
         \midrule
         ALCE \citep{gao:alce}, SuRe \citep{kim2024sure} \& RECOMP \cite{xu2024recomp} & Explore summarization of  retrieved items for compression.
         \\
         Teach LLMs to Personalize \citep{Li2023TeachLT}
         & Context independent and dependent summarization to emphasize key retrieved aspects. \\
         QADecontext \citep{newman-etal-2023-question} $\dagger$
         & Decontextualization as a downstream task when presenting passages from scientific documents. \\
         \midrule
         \midrule
         \multicolumn{2}{c}{Compose (\S\ref{sec:pres_compose})} \\
         \midrule
         Fixed Chunking \citep{Wu2021RecursivelySB} & Recursively summarizes adjacent chunks in books. \\
         RAPTOR \citep{Sarthi2024RAPTORRA}
         & First clusters then summarizes related chunks of text. \\
         \midrule
         \midrule
         \multicolumn{2}{c}{Truncate (\S\ref{sec:pres_truncate})} \\
         \midrule
         FiD-Light \citep{FiD-Light}
         & Extracts vector subsets to speed up attention bottlenecks in FiD-like model decoding.
         \\
         Choppy \citep{Bahri2020ChoppyCT} $\dagger$
         & A supervised approach to ranked list truncation. \\
         \bottomrule
    \end{tabular}}
    \caption{Instances of Presentation-related research. $\dagger$: Relevant for future REML research.}
    \label{tab:presentation}
\end{table}

\subsection{Consumption}
In REML, the predictive model is presented with one or more documents. The effectiveness of $\mlmodel{\mlparam}$ is influenced by the \textit{consumption} of the presented documents. 
Ideally, $\mlmodel{\mlparam}$ would consume all the documents simultaneously, yet our systems are computationally limited; hence, the \textit{granularity} of consumption is typically limited to a subset of the presented documents. Depending on the task, different \textit{consumption algorithms} may prove varying in utility---some algorithms are used for extraction and others for on-the-fly updates of the predictive model parameters. In contrast, \textit{decoding algorithms}, such as beam search \cite{freitag-al-onaizan-2017-beam} or nucleus sampling \cite{Holtzman2020nucleus}, provide ways to decode effective outputs given the presented documents and can incorporate verification for improved effectiveness. There are additional concerns during consumption, such as efficiency \cite{lumen, FiD-Light} and attribution \cite{gao:alce, Schuster2023SEMQASM, asai2024selfrag, menick:deepmind-gophercite}, that provide further utility.
Table \ref{tab:consumption} summarizes the research related to consumption.

\subsubsection{Consumption at different granularities of retrieval items\label{sec:con_gran}} Typically, multiple items are retrieved and it is a design choice whether to process retrieved items separately or together. The followings are the main paradigms for consuming retrieval items:

\begin{itemize}
    \item Single: Only a single item is incorporated in the prediction. This may be sufficient for simple queries, but often information will need to be combined across multiple retrieved items.
    \item Ensemble: Predictions are made for multiple retrieval items in the \textit{single}-fashion, then aggregated \cite{Khandelwal2020Generalization, Shi2023REPLUGRB}.
    \item Joint: When the prediction has a sufficiently flexible context representation, then multiple retrieval items can be passed simultaneously in a single inference procedure \citep{izacard-grave-2021-leveraging, Lewis+al:2020}. This is potentially richer than the ensemble approach since each retrieved item is \textit{aware} of the others. Due to computational limits, only a few retrieval items are able to be processed this way and the \textit{ensemble}-approach is relatively more scalable.
    \item Multi-round: A hybrid approach where a subset of retrieved items are processed at a time, and the next subset incorporates information about the retrieved items processed thus far \citep{jiang2023active-flare}. Although more scalable than the \textit{joint}-approach, this may be slower. That being said, some applications (e.g. dialogue) naturally conform to the \textit{multi-round} framework.
\end{itemize}

These paradigms are atomic functions that can be composed to form more complex operations. 
For instance, given lists of retrieved items, denoted as $r_0, r_1, r_2, r_3$, the atomic functions with different granularities can be composed as:
\begin{equation}
    y = ensemble(multiround(joint(r_0), joint(r_1)), multiround(joint(r_2), joint(r_3)))\text{,}
\end{equation}

\subsubsection{Consumption algorithms\label{sec:con_alg}} Independent of the choice of granularity, there are algorithms applicable for consumption. Broadly, they fall into the following categories:

\begin{itemize}
    \item Extractive: The predictive model is limited to extracting exact information from the retrieved item, e.g., a span of text from a retrieve passage to answer a question \citep{Khandelwal2020Generalization, lan2023copy, wang2023filco}. This can be achieved through pointer networks \citep{NIPS2015_29921001}, constrained decoding \citep{hokamp-liu-2017-lexically, hu-etal-2019-improved, post-vilar-2018-fast}, and other similar techniques \cite{kim-etal-2023-tree}.
    \item Analogical: Learning by example, case based reasoning, and retrieve-and-edit approaches all fall under the category of analogical reasoning. Each of these involves different underlying mechanisms, but essentially the predictive model will be extrapolating from one or more demonstrative examples to make its prediction and not necessarily extracting factual knowledge from the retrieved items \citep{das2020a}.
    \item Contextual: The predictive model incorporates the retrieved items in its context, but the decoding of the output is not constrained in any way \citep{Shi2023REPLUGRB}.
    \item Latent: A hybrid approach where retrieved items are not incorporated directly into the context, but are instead incorporated in other ways, e.g., by merging hidden states \citep{yogatama-adaptive-semiparametric}. Similar to the \textit{contextual}-approach, decoding is not constrained.
\end{itemize}

\subsubsection{Decoding algorithms\label{sec:con_dec}} Independent of the \textit{consumption algorithm}, there are different \textit{decoding algorithms} that can be used for producing outputs in REML. The decoding algorithms explored thus far fall into the following categories:

\begin{itemize}
    \item Output-only: Search algorithms like beam search \citep{freitag-al-onaizan-2017-beam} will only consider the model output when scoring candidate predictions \citep{Khandelwal2020Generalization}.
    \item Retrieval-enhanced: Search algorithms like beam search will consider both the model output and the retrieved items when scoring candidate predictions \citep{Lewis+al:2020, asai2024selfrag}. This should penalize spurious associations between query and retrieved item.
    \item Verification-based: The initial output of the predictive model will be scrutinized by a verification module and potentially rejected if a condition is a met, e.g., the output does not entail the retrieved item \citep{jiang2023active-flare}.
\end{itemize}

\subsubsection{Consumption efficiency\label{sec:con_eff}} Steps can be taken in \textit{presentation} to speed up inference, such as by compressing passages through summarization or truncating the list of retrieved items \cite{lumen, FiD-Light}. There are other approaches to improve efficiency that are more tightly integrated with consumption, e.g., partially precomputing passage embeddings.

\subsubsection{Attribution and other extensions\label{sec:con_att}} Advanced applications of REML will augment the predictive model output space to incorporate REML-specific information. The most common instance of this is probably to support attribution, so that each part of the output can be traced back to the relevant retrieved item \citep{gao:alce, Schuster2023SEMQASM, asai2024selfrag, menick:deepmind-gophercite}. Other cases involve in-line verification or calls to external tools that would not be easily possible without incorporating retrieved items \citep{asai2024selfrag}.

\begin{table}[t]
    \centering
    \adjustbox{max width=\textwidth}{\begin{tabular}{l|l|p{11cm}}
    \toprule
        \toprule
        \multicolumn{3}{c}{Consumption Granularities (\S\ref{sec:con_gran})} \\
        \midrule
        $k$NN-LM \cite{Khandelwal2020Generalization} 
        & Single + Ensemble 
        & Probabilities are computed for each retrieved item independently then combined. \\
        RePLUG \citep{Shi2023REPLUGRB}
        & Single + Ensemble
        & Probabilities are computed for each retrieved item independently then combined. \\
        RAG \citep{Lewis+al:2020} & Joint
        & Retrieved items are concatenated before consumption. \\
        FiD \citep{izacard-grave-2021-leveraging} & Joint
        & Retrieved items are concatenated in the decoder during consumption. \\
        FLARE \citep{jiang2023active-flare}
        & Multi-round
        & Potentially retrieves new items as generation progresses. \\
         \midrule
        \midrule
        \multicolumn{3}{c}{Consumption Algorithms (\S\ref{sec:con_alg})} \\
        \midrule
        kNN-LM \citep{Khandelwal2020Generalization} & Extractive & A single word is selected from the retrieved context. \\
        CoG \citep{lan2023copy}
        & Extractive & An extension of kNN-LM that can output both words and phrases. \\
        FilCo \citep{wang2023filco}
        & Extractive & Filters out irrelevant texts from retrieved documents.  \\
        CBR \citep{das2020a}
        & Analogical & Incorporate knowledge graphs into neural models in spirt of case based reasoning. \\
        Dynamic L2M \citep{Drozdov2022CompositionalSP} & Analogical & Retrieves demonstrations for few-shot prompting. \\
        RePLUG \citep{Shi2023REPLUGRB}
        & Contextual & Uses retrieved items in the context. \\
        SPALM \citep{yogatama-adaptive-semiparametric}
        & Latent & Incorporates retrieved items into the hidden state. \\
         \midrule
         \midrule
         \multicolumn{3}{c}{Decoding Algorithms (\S\ref{sec:con_dec})} \\
         \midrule
         kNN-LM \citep{Khandelwal2020Generalization}
         & Text-only
         & Retrieval probability is ignored for next word prediction. \\
         RAG \citep{Lewis+al:2020}
         & Retrieval-enhanced
         & Retrieval probability is incorporated in beam search. \\
         Self-RAG \citep{asai2024selfrag}
         & Retrieval-enhanced
         & Critic probability is incorporated in beam search. \\
         FLARE \citep{jiang2023active-flare}
         & Verification-based
         & Discards low confidence continuations, triggering retrieval. \\
         \midrule
         \midrule
         \multicolumn{3}{c}{Consumption Efficiency (\S\ref{sec:con_eff})} \\
         \midrule
         LUMEN \cite{lumen} 
         & Precompute
         & Partially computes passage representations offline. \\
         \midrule
         \midrule
         \multicolumn{3}{c}{Attribution and Extensions (\S\ref{sec:con_att})} \\
         \midrule
         ALCE \citep{gao:alce}
         & Attribution & End of sentence. Multi-source. \\
         SemQA \citep{Schuster2023SEMQASM}
         & Attribution & Mid-sentence. Multi-source. \\
         Self-RAG \citep{asai2024selfrag}
         & Attribution & End of sentence. Single source. \\
         Self-RAG \citep{asai2024selfrag} 
         & Verification & Outputs a special token indicating whether to use document. \\
         \bottomrule
    \end{tabular}}
    \caption{Instances of consumption research.}
    \label{tab:consumption}
    \vspace{-15pt}
\end{table}

\section{Storing}
\label{sec:storing}

Storing is one of the optional yet crucial properties of REML models and refers to how retrievable items are saved, represented, and indexed.
The storage components can be categorized into coupled and decoupled storage. 
If at least one external memory is optimized jointly with the predictive model, we call the architecture has \textit{coupled storage}. In the coupled storage architecture, contents can be populated to the storage \textit{online}, updated alongside a predictive model, e.g., Neural Turing Machines (NTM) \cite{graves2014neural, graves2016hybrid} and REALM \cite{guu-realm}.
On the other hand, if all external storage are from off-the-shelf system, and the contents are populated \textit{offline}, we call the architecture has \textit{decoupled storage}, e.g., kNN-LM \cite{Khandelwal2020Generalization} and ED2LM \cite{hui-etal-2022-ed2lm}.

In its simplest form, REML systems will operate with \textit{decoupled} storage where the retrieval model is implemented independent of the predictive model. Since entries in the storage are populated offline and many off-the-shelf retrieval models are readily available, it is relatively convenient to construct the decoupled storage.
On the other hand, many advanced REML systems operate with \textit{coupled} storage where the retrieval model is directly influenced by the predictive model. Here, the entries in the storage are populated online and updated alongside the predictive models. In this section, we describe storage operations and challenges associated with the coupled and decoupled storage.

\subsection{Primary Storage Operations}

In general, there are three types of operations that each storage system must support to be effectively utilized in REML systems: \textit{Address Generation}, \textit{Read}, and \textit{write}. 
The storage operations will typically be used in three scenarios: 1) $\mlmodel{\mlparam}$ needs to incorporate historical context in its prediction, e.g., long context language models \cite{bertsch2023unlimiformer};  2) $\mlmodel{\mlparam}$ conducts various types of online learning using recent/past interactions, e.g., experience replay in reinforcement learning and language agents \cite{shinn2023reflexion}; and 3) $\mlmodel{\mlparam}$ is a memory network-like architecture \cite{weston2015memory-networks} where retrieval is an abstract process and everything written to or read from the storage is controlled by the network which is the service of the objective being optimized.

\subsubsection{Address Generation}

An important aspect of utilizing storage in REML systems is the ability to store and retrieve specific pieces of information. Therefore, the storage system must be capable of generating a specific address for reading from or writing to the storage. Storage location can be divided into abstract location (slots in the storage space) and physical location (where in hardware the storage sits). 
For abstract location, it boils down to the role of $\textit{address}_\textit{location}$, as introduced in Section \ref{subsec:search-eq}. In most cases, it will be a simple rotational function \cite{graves2014neural, graves2016hybrid} which allows an iteration through a sequence of slots. For better efficiency, the function can store entries into clusters by content \cite{weston2015memory-networks} or layers \cite{guillaum-19-large-memory-product-keys}, reducing the computation during searching.
For physical location, entries that do not need to be in RAM or VRAM can be moved to a disk with memory mapping, while entries that must be in the RAM, such as embedding index to be searched, can be compressed without notable performance degradation \cite{izacard_few-shot_2022}.

\subsubsection{Read}
Reading from the storage is an essential part of REML models and closely tied with the search operations.
However in Section \ref{sec:storing}, we focus on how retrievable items are represented in the storage. How they are read are discussed in Section \ref{sec:searching}.

\subsubsection{Write}

At time $t$, after the address vector $w_t$ is obtained, storage operation can be done by a \textit{write} function that updates the datastore $\collection_t$ as follows:
\begin{equation}
    \collection_{t+1} = \textit{write}(w_t, \collection_t, \textit{payload}_t)
\end{equation}

\noindent
where \textit{payload} can be a form of vector or raw representation, which can be preprocessed by some functions, similar to the functions defined in Section \ref{sec:presentation}, before being stored in the storage \cite{hui-etal-2022-ed2lm}.

With the view of addressing mechanism mentioned in Section \ref{subsec:search-eq}, it can be understood that location-based addressing is used before the write operation, following the framework of Neural Turing Machine (NTM) \cite{graves2014neural, graves2016hybrid}.
In most cases, the write operation will simply append the latest entry to the end of a storage (address vector pointing to the next available slot), executed after every new input in sequential order, following the work of neural cache model \cite{grave2017continuous-cache}. 
On the other hand, some architectures, such as Memory Neural Network (MemNN) \cite{weston2015memory-networks} or large memory layers \cite{guillaum-19-large-memory-product-keys} operate differently, and will generate an address specifying where in memory to write the new entry, potentially overwriting any entry that was there before. 
Regularization can be applied in memory networks to ensure that a substantial portion of the memory is used. These decisions can be made by concerted effort with the storage management component.

\subsection{Phases of Storage Operation}\label{subsec:storage-phase}

In REML systems, storage operates through two distinct phases. The initial phase, called \textit{Storage Construction}, involves the system setting up the storage infrastructure with the necessary information to facilitate its operations. Following this setup, the system transitions to the \textit{Storage Management} phase, where it determines the appropriate strategies for storing information, including the selection of what data to retain, the optimal storage locations, and the methods for organizing the information for future retrieval.

\subsubsection{Storage Construction}
REML systems mostly initialize its storage by processing an entire retrieval corpus. This can be done offline before training, after training before inference, and throughout training as needed. 
Storage construction is well documented and studied, and the initial storage construction is essentially a series of write and address generation operations. 
Storage can be constructed as a key-value structure where retrieval space $\docSpace$ can be defined as:
\begin{equation}\label{eq:storage-decoupled-datastore}
    \docSpace = \{(k_i, v_i) \,|\, d \in\collection, \, k_i = transform_k(d), v_i = transform_v(d)\}
\end{equation}
\noindent
where $transform_k$ is a key representation function that can take each entry in the corpus or an input instance $\instance$. Similarly, $transform_v$ is a value representation function that can take each entry in the corpus or an input instance $\instance$ to generate a value in the storage. Note that the $transform_k$ and $transform_v$ function can simply be the identity function, meaning it does not change the key and value at all. 
Table \ref{tab:storage} describes how each paper constructed its storage offline and/or online.
For example, in EMAT \cite{wu-etal-2022-efficient}, where the collection is pairs of questions and answers,  the $transform_k$ function utilizes only the question in each pair to generate the key representation, and $transform_v$ function uses only the answer from each pair to generate the value representation.

\subsubsection{Storage Management}
Once the storage is initialized, both for optimal task completion and efficiency, there is a need to schedule the storage operations, and we dub this phase as storage management. Efficient usage of storage can be understood in terms of space and speed which come down to: when, what, and how to store.

\paragraph{When to store}
In many scenarios, $\rmodel{\rparam}$ will pull information from the storage built at an initial phase based on the need of $\mlmodel{\mlparam}$.
However, this setup can introduce \textit{storage staleness} problem when $\rparam$ changes during $\rmodel{\rparam}$ is being sequentially or jointly trained with $\mlmodel{\mlparam}$. Since $\docKeySpace$ (retrieval space of keys in $\collection$) is constructed by $\rparam$ during the training, it is necessary to refresh the storage as $\rparam$ is updated; this was even mentioned from MemNN \cite{weston2015memory-networks}.
This comes down to asking two questions: 1) \textit{how often to update}, and 2) \textit{what portion of storage to update}.
For the first question, one can either synchronously (every training step) or asynchronously (every $T$ training step) update the storage \cite{retrieval-lm-acl23-tutorial}.
For the second question, one can choose to update the entire storage, a subset of the storage, or refrain from updating the storage at all.

\paragraph{What to store}
Continued from when to store, if the storage are being periodically updated, it may be beneficial to selectively store.
For the synchronous update, updating a subset of storage either by in-batch approximation or reranking is preferred due to large computational overhead, while full storage update is often performed when update is done asynchronously \cite{izacard_few-shot_2022, retrieval-lm-acl23-tutorial, zhong-etal-2022-training, guu-realm}.
Another way to selectively store is to erase some of the entries stored in the past along with the periodic update as storage can become full. 
One simple approach is to set a window size of a storage and manage it like a queue \cite{shinn2023reflexion, wu2022memvit, dai-2019-transformer-xl, Rae2020Compressive}, similar to discarding the oldest entry in the storage \cite{grave2017continuous-cache}. \citet{weston2015memory-networks} devised a separate erasure module that scores the utility of each entry to discard the least useful entries.

\paragraph{How to store}
This encompasses entry representation, e.g., index compression \cite{wu2022memvit, Rae2020Compressive, martins2022-infinite-former} and quantization \cite{izacard_few-shot_2022} and architectural choice of the storage, e.g., key-value structure \cite{grave2017continuous-cache, Khandelwal2020Generalization, wu2022memorizingtransformers, zhong-etal-2022-training, min2023npm, borgeaud:retro, yogatama-adaptive-semiparametric, alon2022neurosymbolic, hui-etal-2022-ed2lm}, where the compression and representation computations can happen incrementally by batch \cite{zamani:reml}.

\subsection{Storage Types}

\begin{table}[t]
\centering
\adjustbox{max width=\textwidth}{\begin{tabular}{lll}
\toprule
\toprule
\multicolumn{3}{c}{Coupled Storage (\S\ref{subsubsec:coupled-storage})} \\
\midrule
          & Key & Value \\
\midrule
NTM \cite{graves2014neural, graves2016hybrid} & transformed output of $\mlmodel{\mlparam}$  & (the same as the Key) \\
MemNN \cite{weston2015memory-networks}, MemN2N \cite{sukhbaatar2015end} \& DMN \cite{kumar2016dynamicmemory} & input feature embedding & (the same as the Key) \\
Neural Cache Model \cite{grave2017continuous-cache} & hidden representation of RNN & next word \\
RUM \cite{seqrec-user-memnet2018} & user-item embedding & (the same as the Key) \\
Transformer-XL \cite{dai-2019-transformer-xl} & hidden representation of Transformer & (the same as the Key) \\
LongMem \cite{wang2023-longtermmem} \& Memorizing Transformer \cite{wu2022memorizingtransformers} & attention-key & attention-value \\
MemTransformer \cite{burtsev2021memtransformer} &  sequence of tokens & tokens \\
RPT \cite{rubin2023longrange-self-retrieval} \& Unlimiformer \cite{bertsch2023unlimiformer} & token chunk embedding & (the same as the Key) \\
MeMViT \cite{wu2022memvit} \& PFMN \cite{lee2018memory360videos} & image frame embedding & (the same as the Key) \\
REALM \cite{guu-realm}, REPLUG LSR \cite{Shi2023REPLUGRB} \& ATLAS \cite{izacard_few-shot_2022} & document embedding & (the same as the Key) \\
EMAT \cite{wu-etal-2022-efficient} & question embedding & answer embedding \\
QAMAT \cite{qamat} & question embedding & question-answer embedding \\
TRIME \cite{zhong-etal-2022-training}  & context (sequence of tokens) & next token \\
NPM \cite{min2023npm} & token embedding & token \\
Reflexion \cite{shinn2023reflexion}, CLIN \cite{majumder2023clin} \& ExpeL \cite{zhao2023expel}  & self-reflection in NL & (the same as the Key) \\
Generative Agent \cite{park2023generative-agents}  & stream of experience in NL & (the same as the Key) \\
Voyager \cite{wang2023voyager}  & program description embedding & program code\\
MemPrompt \cite{madaan-etal-2022-memprompt} & NL question & NL human feedback \\
Synapse \cite{zheng2024synapse} & task metadata embedding & state and trajectories in NL \\
\midrule
\midrule
\multicolumn{3}{c}{Decoupled Storage (\S\ref{subsubsec:decoupled-storage})} \\
\midrule
kNN-LM \cite{Khandelwal2020Generalization} \& SPALM \cite{yogatama-adaptive-semiparametric} & context embedding & next token \\
RAG \cite{Lewis+al:2020} & document embedding & (the same as the Key) \\
RETOMATON \cite{alon2022neurosymbolic} & context embedding & next token, pointer \\
KIF \cite{kif} & evidence embedding (multimodal) & (the same as the Key)  \\
RETRO \cite{borgeaud:retro} & token chunk embedding & token chunk\\
ED2LM \cite{hui-etal-2022-ed2lm} & document embedding & document\\
REPLUG \cite{Shi2023REPLUGRB} & document embedding & (the same as the Key)\\
Embodied-RAG \cite{xie2024embodiedrag} & node of a topological map in NL & position and path to the location \\
\bottomrule
\end{tabular}}
\caption{Instances of REML models with external storage. NL is an abbreviation of natural language.}
\label{tab:storage}
\vspace{-0.5cm}
\end{table}

In the literature, two types of storage architectures are identified in REML systems: \textit{Coupled Storage} and \textit{Decoupled Storage}. The following sections and Table \ref{tab:storage} will elaborate on these architectures.

\subsubsection{Coupled Storage}\label{subsubsec:coupled-storage}
Coupled storage is defined as a storage that can be updated online during training and inference of the predictive model and can be jointly optimized.
Initial developments of the coupled storage enhancements were primarily led by Neural Turing Machine (NTM) \cite{graves2014neural, graves2016hybrid} and Memory Network (MemNN) \cite{weston2015memory-networks}. They showed abstract operations, such as copying, recall, and sorting, to language reasoning tasks by leveraging external addressable storage, where contents in the storage of these architectures are dense vectors which can be readily used by $\mlmodel{\mlparam}$. 
We refer the readers to the original papers of the two models for deeper understanding of primary shape of REML models with coupled storage.
There are a few notable characteristics about REML with coupled memory, including but not limited to the following.

\paragraph{Staleness of Coupled Storage}
One of the primary concerns of the coupled storage arises when $\rmodel{\rparam}$ is being trained, making the storage stale.
It is still an ongoing challenge in the community, but as mentioned in Section \ref{subsec:storage-phase}, there have been several techniques devised to circumvent this issue by answering \textit{how often to update} (synchronous or asynchronous) and \textit{what portion of the storage to update} (full or partial). Therefore, there are five different strategies including an avoidance of update. 
\textbf{Synchronous full update}
is the simplistic approach to solve the staleness problem by updating the storage at every training step. It is attempted in a few research \cite{rubin2023longrange-self-retrieval, bertsch2023unlimiformer}, but its large computational overhead prevents it from being used in a practical setting \cite{izacard_few-shot_2022}.
\textbf{Synchronous partial update} can be done by selecting a batch of entries to update \cite{izacard_few-shot_2022}. Depending on the applications, there can be various batch selection strategies, such as lexical similarity \cite{zhong-etal-2022-training} or in-document sampling \cite{min2023npm}.
\textbf{Asynchronous full update} is done by updating the full storage every $T$ training steps \cite{guu-realm, izacard_few-shot_2022, Shi2023REPLUGRB, wu-etal-2022-efficient}. This allows staleness in the index before it is updated again. For example, \citet{wu-etal-2022-efficient} freeze the storage at the beginning of each training epoch and only updates at the end of each epoch.
As far as we know, there is little attempt on \textbf{asynchronous partial update} as it may degrade the training performance in larger margin.
Alternatively, since index recreation is highly expensive, it is possible to \textbf{ignore the staleness} and avoid re-indexing with adequate strategies without compromising a performance in a large margin \cite{Rae2016-scaling-ntm, Lewis+al:2020, izacard_few-shot_2022, guu-realm, wang2023-longtermmem, wu2022memorizingtransformers}.

\paragraph{Cold Start Problem in Coupled Storage}
Another characteristic of the models equipped with coupled storage is that they can have a cold start problem, where the performance of $\mlmodel{\mlparam}$ is suboptimal before the storage is filled up with enough information. Most of the architectures that start with an empty storage such as language agents (LA) or long-context language models that process a long document in multiple training steps have this issue \cite{park2023generative-agents, majumder2023clin, shinn2023reflexion, zhao2023expel, wang2023voyager, wu2022memorizingtransformers, zhong-etal-2022-training}.
However, the cold start problem can be alleviated when the model can be adapted to a new task and the storage/experience built during the previous tasks is transferable to the new task \cite{majumder2023clin}.

\paragraph{Versatility of Coupled Storage}
Despite its disadvantages, coupled storage has seen significant development both theoretically and in various applications.
It includes making REML systems with coupled storage end-to-end trainable \cite{sukhbaatar2015end}, capturing the position and temporality of language using episodic storage \cite{kumar2016dynamicmemory}, and scaling the storage through sparse access \cite{Rae2016-scaling-ntm, zaremba2016rlntm} and strategic storage management \cite{guillaum-19-large-memory-product-keys, grave2017continuous-cache}.
These have led to applications in various domains, such as meta-learning \cite{metalearning-mann}, sequential recommendation \cite{seqrec-user-memnet2018}, video summarization and recognition \cite{lee2018memory360videos, wu2022memvit}.
In long context language modeling \cite{bertsch2023unlimiformer, zhong-etal-2022-training}, there have been approaches to solve the task through attention recurrence \cite{dai-2019-transformer-xl, yogatama-adaptive-semiparametric, wu2022memorizingtransformers, wang2023-longtermmem}
and compression of hidden states \cite{Rae2020Compressive, martins2022-infinite-former, wu2022memvit}.
%
Recently, research on language agents focus on agents' ability to use language models for perceiving, reasoning, planning, and managing memory while interacting with external environments \cite{sumers2023cognitive-CoALA}. These agents, which learn independently from their observations and adapt their knowledge, are equipped with external storage for long-term memory, allowing them to store past reasoning \cite{majumder2023clin, wang2023voyager, zhao2023expel, park2023generative-agents, shinn2023reflexion} or user feedback \cite{madaan-etal-2022-memprompt} for future use and self-reflection \cite{madaan2023selfrefine}.

\subsubsection{Decoupled Storage}\label{subsubsec:decoupled-storage}
Decoupled storage is defined as a storage method in REML systems where $\rmodel{\rparam}$ operates independent to $\mlmodel{\mlparam}$. In contrast to the coupled storage, where the entries are being updated dynamically online usually influenced by joint optimization of $\mlmodel{\mlparam}$ and $\rmodel{\rparam}$, decoupled storage involves offline population of entries, i.e., storage becomes read-only during the online stage. There are a few notable characteristics about REML systems with decoupled memory, including but not limited to the following.

\paragraph{Ease of Implementation.}
Since the retriever is completely independent from the predictive model, the implementation of the REML system becomes a lot easier compared to that with coupled storage \cite{borgeaud:retro, hui-etal-2022-ed2lm, yogatama-adaptive-semiparametric, Khandelwal2020Generalization, alon2022neurosymbolic}. When training the REML system, in decoupled storage architecture, one can either train the retriever separately or use an off-the-shelf retriever already publicly available \cite{Shi2023REPLUGRB}. 
This also means that one can easily edit a REML system by simply replacing its components.
Therefore, this design can guarantee a liberation from storage staleness and cold start problem unlike the coupled storage architecture.
The ease of implementation stands out when the systems need to incorporate multiple storage that are multi-modal or multi-source \cite{kif, yang2023kg-reml-survey, knowledge-enhanced-nlg-survey}, where it can be tricky to be implemented with coupled memory architecture.

\paragraph{Performance sub-optimality.} Generally, it is known that in REML systems with multiple components, end-to-end training yields a better performance compared to training each component individually \cite{sachan-etal-2021-end, wang2024llm, zamani2024stochastic}. However, in REML systems with decoupled storage, the storage component and predictive model are trained separately. This configuration might lead to sub-optimal performance in the system's downstream tasks. In other words, despite the convenience, the fixed nature of the storage during the training of the predictive model, and vice versa, in decoupled storage systems, prevents them from adapting to each other's needs.

\section{Optimization}
\label{sec:optimization}

As depicted in Figure~\ref{fig:reml-framework}, a REML system consists of the multiple components interacting with each other. Optimization can be done either end-to-end (i.e., optimizing all model parameters simultaneously for a common goal). Alternatively, optimization can be done for a subset of model parameters, such as independent optimization of each component in REML. For instance, for optimizing the Query Generation component, ground-truth queries are required for independent optimization of this component. Obtaining such data is difficult to obtain for some components. Distant or weak supervision is a potential solution to address this issue. In the rest of this section, we mainly focus on the optimization of the retrieval model and the predictive model as the two main components of REML systems.

\subsection{Retrieval Model Optimization}

\subsubsection{No REML Optimization} 
In a wide range of studies, retrieval models are not optimized. In many of them, queries and documents are in the form of unstructured text. In that case, query and document representations are often computed based on term statistics, such as term frequencies in the documents or document frequencies in the document collection. Using retrieval models such as TF-IDF \cite{Salton1988TFIDF}, BM25 \cite{Robertson1995OkapiBM25}, and query likelihood \cite{Ponte:1998}, with default parameters, belongs to this category. For example. the Dr.QA model \cite{chen-etal-2017-reading} uses the ElasticSearch implementation of TF-IDF for document retrieval from Wikipedia for factoid question answering. The SelfMem model \cite{selfmem} uses BM25 for document retrieval for a number of retrieval-augmented text generation tasks, such as translation and dialogue.

Pre-trained language models can be also used to produce latent representations for queries and documents, where simple similarity functions, such as dot product or cosine similarity, are used for computing relevance scores for a query-document pair. Even though these language models went through expensive optimization procedures, their optimizations are not REML-specific. Note that employing language model representations for retrieval with no optimization often do not perform well. For instance, \citet{Lien:2023:GWS} demonstrated that using plain BERT or RoBERTa representations for zero-shot retrieval is substantially worse than term matching models, like BM25. 

More recently, it has been observed that large-scale instruction-tuned language models, such as GPT-3.5, can be carefully instructed to rank a few documents for a given query \cite{Sun:2023:InContextLTR}. These models can perform effectively, but cannot be applied to large document collections and could be only used as re-ranking models. 

Retrieving from databases through structured queries, such as SQL, also belong to this category. A wide range of task-oriented dialogue systems, such as intelligent assistants for travel booking and restaurant reservation, require access to databases for up-to-date availability \cite{DialogueSurvey}.

\subsubsection{Independent Optimization}
REML may take advantage of retrieval models whose optimization is independent of the predictive model's parameters. Retrieval models are often trained using a set of \texttt{query-document-relevance} triplets. The \texttt{relevance} signal may come from (1) explicit annotations, e.g., from expert assessors or crowdworkers, (2) implicit feedback \cite{Joachims:2002:Click,Joachims:2017:ULTR}, e.g., user clicks, dwell time, mouse movements, etc, or (3) automatically generated weak signals (also known as distant supervision signals), such as appearance of a phrase, e.g., answer in the context of QA, in the documents, retrieval scores from another retrieval model \cite{Dehghani2017Weak}, or annotations produced by large language models \cite{thomas2023large}.
Using these training triplets, retrieval models can be optimized using three different approaches: (1) pointwise, (2) pairwise, or (3) listwise ranking. Refer to \citet{Liu2009LTR} for more information on various ranking loss functions.

In the context of REML, a large set of studies, e.g., \cite{vu2023freshllms,Hashemi:2021:MultipleRep,lyu:2023-data-importance,jiang2023active-flare}, use commercial search engines, such as Bing or Google, as their retrieval models. These search engines are optimized using a combination of the relevance signals mentioned above. Therefore, they are considered as independent optimization models. A set of approaches, such as \cite{izacard-grave-2021-leveraging}, train retrieval models on explicitly labeled collections, such as MS MARCO \cite{Campos2016MSMARCO} or provenance labels in the KILT benchmark \cite{petroni-etal-2021-kilt}, and then use the trained models on an often out-of-domain REML scenario. Weak or distant supervision is also used in \cite{qu-2021-multimodal-passage-retrieval,Qu:2020:OR-ConvQA,Salemi2023PreTrainingMD} for open-domain (visual) question answering by assuming that any document that contains the answer phrase is relevant.

\subsubsection{Conditional Optimization}
In conditional optimization, the retrieval model is optimized, conditioned on the predictive model $f_\theta$. A group of conditional optimization approaches use knowledge distillation. For instance, \citet{izacard2021distilling} use the aggregation of cross-attention weights in the fusion-in-decoder architecture as weak signals to train the retrieval model. Here, the decoder that provides the weights plays the role of a teacher model and a dense passage retrieval model plays the role of a student model. Alternatively, \citet{DBLP:journals/corr/abs-2010-10999} use the similarity score produced by an answer span selection model, i.e., the reader, as teacher scores and minimize the KL-divergence between them and retrieval scores. 

As shown by \citet{izacard2021distilling}, knowledge distillation from the downstream ML model to the retrieval model can be done iteratively, as follows:
\begin{equation}
    \omega^{(t+1)}=\arg \min _\omega \frac{1}{|T|} \sum_{(\instance, \target) \in T} \loss\left(\mlmodel{\theta^{(t)}}\left(\instance ; \rmodel{\omega}\right), \target\right)
\end{equation}
where the retrieval model in iteration (or epoch) $t+1$ is optimized based on the parameters of the predictive model at iteration $t$, where $\mathcal{L}$ is the downstream loss function.

\subsection{Predictive Model Optimization}
\subsubsection{No REML Optimization} 
Similar to retrieval models, predictive models may also be used as a `black-box' systems without REML-specific training. For instance, a wide range of query expansion approaches, such as the Rocchio's algorithm \cite{Rocchio:1971}, relevance models \cite{Lavrenko:2001:RM}, and divergence minimization model \cite{Zhai2001Mix} expand the queries based on the appearance of terms and concepts in the retrieval results. Using pre-trained large language models in a zero-shot setting is another example that has received considerable attention in recent years \cite{Shi2023REPLUGRB,salemi:lamp}.

\subsubsection{Independent Optimization}
Predictive models in REML can be optimized independent of the retrieval model's parameters. For instance, we can optimize predictive models by assuming that the retrieval model is optimal (i.e., retrieving ground truth relevant documents). In this case, the optimization of predictive model $f_\theta$ can be modeled as:
\begin{equation}
\theta^*=\arg \min _\theta \frac{1}{|T|} \sum_{(\instance, \target) \in T} \loss\left(\mlmodel{\theta}\left(\instance ; \rmodel{\mathrm{opt}}\right), \target\right)
\end{equation}
where $\mathcal{L}$ is the loss function for the downstream task. For instance, a number of open-domain QA models are optimized to extract or generate answers given the question and the gold (ground truth) passage \cite{chen-etal-2017-reading}. Some may relax the optimality assumption of retrieval models and inject non-relevant documents to the ground truth set. These documents can be either sampled randomly or from the output of a retrieval model, but not $g_\omega$.

\subsubsection{Conditional Optimization}
Alternatively, predictive models can be trained conditioned on the retrieval model's performance. Without loss of generality, this can be seen as an iterative process, where the predictive model is optimized in one iteration and the retrieval model is optionally optimized in the next iteration. With this formulation, the parameters of a predictive model at iteration $t$ can be obtained as follows:
\begin{equation}
\theta^{(t)}=\arg \min _\theta \frac{1}{|T|} \sum_{(\instance, \target) \in T} \loss\left(\mlmodel{\theta}\left(\instance ; \rmodel{\omega^{(t)}}\right), \target\right)
\end{equation}

\subsection{Joint Optimization of Retrieval and Predictive Models}

\subsubsection{Joint Multi-Task Optimization}
Retrieval and predictive models can be trained jointly. Joint optimization can be modeled end-to-end (explained later in this section) or through multi-task learning. In joint multi-task optimization, for any training instance, both the retrieval results and the predictive model parameters will be updated. For instance, FiD-Light \cite{FiD-Light} generates the documents with positive provenance score in addition to the output text for retrieval-augmented text generation tasks. The generated document IDs are then used for re-ranking the result list. Therefore, this can be seen as a joint optimization of re-ranking and generation.

\subsubsection{End-to-End Optimization}
Following the risk minimization framework, end-to-end optimization in REML can be modeled as follows:
\begin{equation}
\theta^* , \omega^* = \arg \min _{\theta,\omega} \frac{1}{|T|} \sum_{(\instance, \target) \in T} \loss\left(\mlmodel{\theta}\left(\instance ; \rmodel{\omega}\right), \target\right)
\end{equation}
where both parameters sets $\theta$ and $\omega$ get updated simultaneously by optimizing an appropriate loss function for the downstream machine learning task. End-to-end optimization of REML, however, can be challenging. It is mostly due to the top $k$ item selection process of information access models in REML that makes the end-to-end REML model non-differentiable. Existing work make some simplifying assumptions to turn the optimization to a differentiable process. For instance, the RAG model from \citet{Lewis+al:2020} by marginalizing the retrieved document set to a set of pre-selected documents. A similar approach was later utilized by \citet{sachan-etal-2021-end} for open-domain question answering. In addition to marginalization, RetGen \cite{RetGen:2021} and EMDR\textsuperscript{2} \cite{NEURIPS2021_da3fde15}  make a document independence assumption and computes the loss function as a summation over each individual document.

\section{Evaluation}
\label{sec:evaluation}

Our goal in evaluation is to understand whether a change to the system---including a full replacement---is better than keeping the status quo.  For example, we might be interested in knowing whether changing the search component improves predictive performance.  We will refer to this evaluation metric as $\metric$, whose arguments will be explained shortly.  We compute the expected metric value with respect to a distribution over some population $\prob(\inputSpace)$, which is ideally the same distribution used for training data.

We classify evaluation as either \textit{extrinsic}, looking at the final performance of the predictive model, or \textit{intrinsic}, looking at the performance of a component of the system using a local measure of quality rather than predictive model performance \cite{ksj:intrinsic-extrinsic}.  An intrinsic evaluation of a model can be an efficient approximation for an extrinsic evaluation or can measure some independent value such as resource consumption.

\subsection{Extrinsic evaluation} 
 In all situations, we are most often interested in the expected value of the metric for a system.  That is, for a model $\jointModel=\langle\mlmodel{\mlparam},\rmodel{\rparam}\rangle$ and evaluation data $\testdata$, compute,
\begin{align}
\expect[\metric(\jointModel(\instance))]&=\frac{1}{|\testdata|}\sum_{x\in\testdata}\metric(\jointModel(\instance))
\end{align} 
When evaluating a system extrinsically, we can pose hypotheses about relative system performance in several ways \cite{guu-realm, petroni-etal-2021-kilt, Lewis+al:2020}. In \textit{non-overlapping system comparison}, given two model tuples $\jointModel=\langle\mlmodel{\mlparam},\rmodel{\rparam}\rangle$ and $\jointModel'=\langle\mlmodel{\mlparam'},\rmodel{\rparam'}\rangle$, determine if $\expect[\metric(\jointModel(\instance))]>\expect[\metric(\jointModel'(\instance))]$. In \textit{fixed retrieval model comparison}, given two model tuples $\jointModel=\langle\mlmodel{\mlparam},\rmodel{\rparam}\rangle$ and $\jointModel'=\langle\mlmodel{\mlparam'},\rmodel{\rparam}\rangle$, determine if $\expect[\metric(\jointModel(\instance))]>\expect[\metric(\jointModel'(\instance))]$. As a special case, we can consider  $\rmodel{\rparam^*}$ an optimal ranker according to some intrinsic criteria; this allows us to examine whether a system can effectively incorporate relevant items. In \textit{fixed predictive model comparison}, given two model tuples $\jointModel=\langle\mlmodel{\mlparam},\rmodel{\rparam}\rangle$ and $\jointModel'=\langle\mlmodel{\mlparam},\rmodel{\rparam'}\rangle$, determine if $\expect[\metric(\jointModel(\instance))]>\expect[\metric(\jointModel'(\instance))]$. In this case, we can consider $\mlmodel{\mlparam^*}$ as an optimal predictive model according to some intrinsic criteria; this allows us to examine whether a system can effectively retrieve relevant items.

\subsection{Intrinsic evaluation}

REML systems comprise numerous components, each capable of individual assessment. Intrinsic evaluation of a component involves comparing systems based on their isolated performance with respect to that component. Nevertheless, such systems' most important components are the retrieval and predictive models. 

\subsubsection{Intrinsic evaluation of  retrieval}

Intrinsic evaluation of a retrieval model focuses on comparing systems according to their isolated retrieval performance.  In this case, assuming single-turn retrieval, we can pose two styles of hypothesis.  In \textit{non-overlapping system comparison}, given two model tuples $\jointModel=\langle\mlmodel{\mlparam},\rmodel{\rparam}\rangle$ and $\jointModel'=\langle\mlmodel{\mlparam'},\rmodel{\rparam'}\rangle$, determine if $\expect[\metric(\rmodel{\rparam}(\mlmodel{\mlparam}(\instance)))]>\expect[\metric(\rmodel{\rparam'}(\mlmodel{\mlparam'}(\instance)))]$, where, with some abuse of notation, $\mlmodel{\mlparam}(\instance)$ and $\mlmodel{\mlparam'}(\instance)$ considers only the query processing for $\rmodel{\rparam}$ and $\rmodel{\rparam'}$.  In \textit{fixed query processing comparison}, given two model tuples $\jointModel=\langle\mlmodel{\mlparam},\rmodel{\rparam}\rangle$ and $\jointModel'=\langle\mlmodel{\mlparam},\rmodel{\rparam'}\rangle$, determine if $\expect[\metric(\rmodel{\rparam}(\mlmodel{\mlparam}(\instance)))]>\expect[\metric(\rmodel{\rparam'}(\mlmodel{\mlparam}(\instance)))]$. 
The choice of metric $\metric$ depends on the task but should be some  retrieval metric, unless the retrieval result is not a ranking.  Such metrics require some relevance estimate for each item.  In the case of REML, this can come from,
\begin{enumerate}
    \item Explicit labels gathered from human raters. This requires instances, targets, and items to be interpretable. 'provenance Labels' in the KILT benchmark for some tasks such as Natural Questions \cite{kwiatkowski-etal-2019-natural} and ELI5 \cite{fan-etal-2019-eli5} can be thought as such labels. 
    \item Inferred labels from the target.  For example, in QA, we could compute the similarity between a retrieved item and the target. `Context Relevance' from RAGAs \cite{es-etal-2024-ragas} and ARES \cite{saadfalcon2023ares} can be thought as a variant of this case.
    \item Attributed labels from the model prediction.  For example, in QA, if a model generates an answer correctly, we can try to attribute the answer correctness to each of the retrieved items. This method, drawing inspiration from eRAG \cite{salemi2024evaluating}, assesses the retrieval model's performance by quantifying the contribution of each retrieved document towards achieving the correct answer. 
\end{enumerate}

\subsubsection{Intrinsic evaluation of  consumption} 
Intrinsic evaluation of consumption focuses on comparing systems according to their isolated ability to translate retrieval results into effective predictions.  Although extrinsic evaluation measures the effectiveness of the system in general, intrinsic evaluation of consumption focuses on whether a prediction is attributable to retrieval results (e.g., versus information already in the consumption model parameters).  In \textit{fixed retrieval comparison}, given two model tuples $\jointModel=\langle\mlmodel{\mlparam},\rmodel{\rparam}\rangle$ and $\jointModel'=\langle\mlmodel{\mlparam'},\rmodel{\rparam}\rangle$, determine if $\expect[\metric(\mlmodel{\mlparam}(\resultSpace_{\rparam,\instance}),\resultSpace_{\rparam,\instance})]>\expect[\metric(\mlmodel{\mlparam'}(\resultSpace_{\rparam,\instance}),\resultSpace_{\rparam,\instance})]$. 
The choice of metric $\metric(\tilde{y},\tilde{\resultSpace})$ depends on the task but measures whether the prediction $\tilde{y}$ is related to the retrieval results $\tilde{\resultSpace}$.
`Faithfulness' from RAGAs \cite{es-etal-2024-ragas} and ARES \cite{saadfalcon2023ares} can be thought as a variant of this case.

\subsection{Datasets and Benchmarks}
\label{subsec:datasets}

REML systems can be evaluated on various tasks. In literature, various benchmarks and datasets serve to assess these systems from diverse angles. Broadly, datasets fall into two categories: 1) those exclusively considering extrinsic evaluation of REML systems, assessing them solely based on end-to-end performance. 2) those furnishing retrieval relevance labels for intrinsic evaluation in addition to end-to-end performance. 
Table~\ref{tab:datasets} illustrates the most commonly employed datasets and benchmarks in the literature.

\begin{table}[t!]
    \centering
    \adjustbox{width=\textwidth}{\begin{tabular}{lll}
        \toprule
        \toprule
        \multicolumn{3}{c}{End-to-End Evaluation (\S\ref{subsec:datasets})} \\
        \midrule
        Task & {Datasets} & Corpus \\
        \midrule
        {Entity Related QA} & PopQA\cite{mallen-etal-2023-when-not-to-trust}, EntityQuestions\cite{sciavolino-etal-2021-simple} & Wikipedia \\
        {Current Events Related QA} & RealtimeQA\cite{kasai2023realtime} & News Websites \\
        {Science Related Multiple-choice QA} & ARC \cite{clark2018think} & Subset of Web \\
        {Science Related QA} & Qasper\cite{dasigi-etal-2021-dataset} & Scientific Articles\\
        Story Related Long-form QA & NarrativeQA\cite{kocisky-etal-2018-narrativeqa} & A Long Story \\
        
        Query-based Summarization & QMSum\cite{zhong-etal-2021-qmsum} & A Meeting Transcript \\

        Personalized Classification and Generation & LaMP\cite{salemi:lamp} & A User Profile \\
        \toprule
        \multicolumn{3}{c}{End-to-End \& Retrieval Evaluation (\S\ref{subsec:datasets})} \\
        \midrule
        {Open-domain Multi-Hop QA} & 2WikiMultiHopQA\cite{ho-etal-2020-constructing}, HotpotQA\cite{yang-etal-2018-hotpotqa, petroni-etal-2021-kilt} & Wikipedia \\
        Open-domain Short-form QA &  Natural Questions\cite{kwiatkowski-etal-2019-natural, petroni-etal-2021-kilt}, TriviaQA\cite{joshi-etal-2017-triviaqa, petroni-etal-2021-kilt}, StrategyQA\cite{geva2021did} & Wikipedia \\
        Open-domain Long-form QA & ELI5\cite{fan-etal-2019-eli5, petroni-etal-2021-kilt}, ASQA\cite{gao:alce} & Wikipedia \\
        Dialogue Generation & Wizard of Wikipedia\cite{dinan2018wizard, petroni-etal-2021-kilt} & Wikipedia \\
        Slot Filling & ZeroShot RE\cite{levy-etal-2017-zero, petroni-etal-2021-kilt}, T-REx\cite{elsahar-etal-2018-rex, petroni-etal-2021-kilt} & Wikipedia \\
        Entity Linking & AIDA CoNLL-YAGO\cite{hoffart-etal-2011-robust, petroni-etal-2021-kilt}, WNED-WIKI/CWEB \cite{10.3233/SW-170273, petroni-etal-2021-kilt} & Wikipedia \\
        Fact Verification & FEVER\cite{thorne-etal-2018-fever, petroni-etal-2021-kilt} & Wikipedia \\
        Open-domain Visual QA & OK-VQA\cite{marino2019okvqa, qu-2021-multimodal-passage-retrieval} & Wikipedia \\
        Open-domain Visual QA & FVQA\cite{wang2017fvqa} & A Supporting Facts Set \\
        
        \midrule
    \end{tabular}}
    \caption{Datasets available for training and evaluating REML systems (not an exhaustive list). Some focus on end-to-end evaluation, while others provide retrieval evaluation labels.}
    \label{tab:datasets}
    \vspace{-0.5cm}
\end{table}

\section{Future Directions}
\label{sec:future}
To enhance REML systems, we propose future work for each of the previously discussed sections.

\subsection{Querying}
\subsubsection{Query with Instruction}
Recent advancements in instruction tuning for LLMs have demonstrated substantial improvements in performance across downstream tasks \cite{Wei2021FinetunedLM, Mishra2021CrossTaskGV, wang-etal-2022-super}. Moreover, recent research on retrieval utilizing instructions has surpassed competitive baselines, showcasing superior performance in terms of retrieval efficiency \cite{asai-etal-2023-task}. With that in mind, developing transformation functions for query generation that produce task and query-specific instructions alongside the query can significantly enhance the retrieval model's capacity to fulfill the requirements of the predictive model. 

\subsubsection{Retrieval System Aware Query Generation}
Most query generation and decomposition functions overlook the type and configuration of the retrieval model, but aligning queries with the model’s specifics can enhance performance. For instance, BM25, which emphasizes exact term matching, performs better with queries that closely align with terms in relevant documents. On the other hand, dense retrieval models, which focus on semantic similarity, benefit more from queries that are semantically aligned with the content of relevant documents. Tailoring query generation to the retrieval model ensures the queries meet the model’s unique requirements, improving retrieval effectiveness.

\subsubsection{Dissociated Interface between Retrieval and Predictive Model}
Most REML systems use natural language for communication between predictive and retrieval models. Models like kNN-LM \cite{Khandelwal2020Generalization}, which uses hidden representations of the predictive model as queries and keys, the representations are all from the predictive model. However, relying solely on natural language or the representation of one model may not be optimal. An alternative is to train both retrieval and predictive models jointly to learn a shared hidden space, enabling more effective communication. This approach can convey information more efficiently and enhance the interaction between the models, leading to better performance and coordination.

\subsection{Searching}
\subsubsection{Predictive Model-Aware Retrieval Systems}
Approaches like search personalization could be used for tailoring retrieval results to a specific predictive model. For example, there can be a situation where multiple predictive models are being served by a few retrieval models \cite{salemi2024search}. In this case, the retrieval model can store meta-data of the predictive models they are serving, opening up opportunities to tailor the retrieval results to each predictive model.

\subsubsection{Redefining Relevancy}
In predictive model-aware retrieval systems, IR models usually evaluate a document's relevance based on how well it meets a human user’s information need. However, in REML, the primary consumer is the predictive model, not the human user. Research shows that \textit{utility judgments}---assessing a document’s usefulness for a specific task---correlate more strongly with the REML model’s performance than traditional relevance labels \cite{salemi2024evaluating, kim2024fairragimpactfair, zhang2024large}. This underscores the need for further research on how to best represent relevance in the context of REML.

\subsection{Presentation \& Consumption}
\subsubsection{Task-Specialized Presentation and Consumption} Similar to how task-specific retrieval is beneficial for distinguishing between fact verification, entity linking, QA, and so on, it will likely be better to use a document representation specific to the task at hand. This may materialize as a prompt with task-specific instructions, task-specific intermediate steps (including explanation for how the document is relevant), or even task-specific embeddings of documents.

\subsubsection{Proactive REML}
In practice, it is beneficial to not only answer the immediate question posed by the query, but also address potential follow up questions \cite{Samarinas_2024_ProCIS, shah23_proactive}. Follow up questions can transition to a new topic (e.g. purchasing a hotel after booking a flight), or dive deep on part of the initial answer.

\subsection{Storing}
\subsubsection{Shared Storage}
Figure \ref{fig:reml-framework} depicts a single $\mlmodel{\mlparam}$ interacting with multiple information access system. However, it is also possible to map multiple $\mlmodel{\mlparam_i}$ to a single information access system. In this scenario, where multiple predictive model is sharing a single collection, the ability of the models to learn what is important to store becomes crucial as pushing irrelevant content to the shared storage can cause degradation in its usefulness and cause performance degradation of other $\mlmodel{\mlparam_i}$ sharing the same collection. This direction is particularly valuable to explore in multi-agent settings \cite{park2023generative-agents}.

\subsubsection{Storage Staleness}
While training the REML model, the updated retriever often makes the previously-built storage stale. Although there have been many attempts to detour this issue, no studies have found a profound way to solve the problem. This persistent challenge necessitates further research into adaptive storage mechanisms that can dynamically align with retriever updates, ensuring data integrity and model efficiency.

\subsection{Optimization}
\subsubsection{Effective and Efficient End-to-End Optimization}
The non-differentiable nature of some components in REML or their interaction makes it challenging to optimize REML systems end-to-end. Existing approaches are based on some simplifying assumptions and developing more accurate and robust approximations for end-to-end REML optimization is an important research direction. Moreover, reward-based optimization of these systems, based on both human and AI feedback, is relatively underexplored. Better understanding of exploration and exploitation of information items provided by the information access system is required.

\subsubsection{Learning from Online and Session-based Feedback}
Interaction between the predictive and information access models can be sequential. Simple forms of such sequential interaction already exists in the context of multi-hop question answering. Using the feedback provided by the predictive model during an inference session and its users to adjust the REML output is critical to develop effective \emph{interactive} REML systems.

\subsubsection{Efficient Approximation of Feedback for Optimization}
Both end-to-end and conditional optimization approaches require feedback from different components of the REML system. As the field moves towards larger and more expensive, developing efficient and accurate feedback approximations could substantially reduce the cost of REML training. This would not only reduces the monetary cost associated with REML training, but would also speed up research progress and lead to more sustainable and environmental-friendly systems.

\subsubsection{One Information Access and Multiple Predictive Models}
Most prior work focuses on developing a REML system for one task. On the other hand, information access systems can serve multiple predictive models, similar to the current search engines that serve billions of users \cite{salemi2024search}. Optimizing information access components that provide service to multiple predictive models, aggregating and calibrating feedback across predictive models, and ``personalizing'' the retrieval result lists for each predictive model are important future directions.

\subsection{Evaluation}
\subsubsection{Broadening Evaluation Approaches}
Evaluations of REML systems predominantly focus on extrinsic measures of system effectiveness, and studies that conduct both extrinsic and intrinsic evaluations mainly employ reference-based methods with relevance labels. Researchers could broaden this scope by assessing additional qualities of REML systems, such as citation quality \cite{gao:alce}, robustness \cite{yoran2024making}, and context utilization \cite{ru2024ragchecker}. Additionally, an often overlooked aspect is the evaluation of trustworthiness and societal impact; although recognized as important, these factors are typically discussed in non-REML contexts like pure information retrieval or text generation. In REML applications, trustworthiness evaluation might present new research directions. For instance, existing definitions of fairness may require revision within the REML context \cite{kim2024fairragimpactfair}, and the propagation of related issues through retrieval and consumption processes may need re-examination. Regarding transparency in evaluation, most model-based, reference-free automatic evaluations of REML models rely on large language models \cite{saadfalcon2023ares, es-etal-2024-ragas, salemi2024evaluating, guinet2024automated}, which often lack transparency because they do not provide clear reasoning for their assessments. These automatic evaluation models can be meta-evaluated based on measurement theory \cite{hand2010measurement} for further validation.

\section{Conclusion}
\label{sec:conclusion}
In this work, we survey the current literature on retrieval-enhanced machine learning (REML) and synthesize it into consistent mathematical concepts, providing researchers with a formalized framework for REML. By bridging the gap between information retrieval research and REML, we also identify new opportunities and open avenues for future studies in this emerging research paradigm.

\bibliographystyle{ACM-Reference-Format}
\bibliography{00-reml}


\end{document}